\pdfoutput=1

\documentclass[11pt]{article}

\usepackage[]{acl}
\usepackage{multirow}
\usepackage{graphicx}
\usepackage{booktabs}
\usepackage{tabularx}
\usepackage{CJKutf8}
\usepackage{amssymb}
\usepackage{amsmath}
\usepackage{tabu} 
\usepackage{hyperref}
\usepackage{breakurl}
\usepackage{times}
\usepackage{latexsym}
\usepackage{tikz}
\usepackage{pgfplots}
\usepackage[normalem]{ulem}
\usepackage{xcolor}
\usepackage{dashrule}
\usepackage{colortbl}
\usepackage{fontawesome}
\usepackage{tipa}
\usepackage{arydshln}

\usetikzlibrary[patterns]
\usepgfplotslibrary{fillbetween}

\definecolor{brickred}{HTML}{b92622}
\definecolor{midnightblue}{HTML}{005c7f}
\definecolor{salmon}{HTML}{f1958d}
\definecolor{burntorange}{HTML}{f19249}
\definecolor{junglegreen}{HTML}{4dae9d}
\definecolor{forestgreen}{HTML}{499c5e}
\definecolor{pinegreen}{HTML}{3d8a75}
\definecolor{seagreen}{HTML}{1EC652}
\definecolor{limegreen}{HTML}{97c65a}
\definecolor{redtable}{HTML}{E67F72}
\definecolor{orangetable}{HTML}{E67F72}
\definecolor{greentable}{HTML}{E67F72}

\usepackage{lipsum}

\usepackage[T1]{fontenc}
\usepackage[utf8]{inputenc}

\usepackage{microtype}

\title{Multi-Task Instruction Tuning of LLaMA for Specific Scenarios: \\ A Preliminary Study on Writing Assistance}

\newcommand{\suda}{\textsuperscript{\faStarO}}
\newcommand{\tencent}{\textsuperscript{\faMoonO}}
\newcommand{\macau}{\textsuperscript{\faSunO}}

\author{
    \textbf{Yue Zhang}\suda \footnotemark[1]\hspace{0.5mm},
    \textbf{Leyang Cui}\tencent,
    \textbf{Deng Cai}\tencent,
    \textbf{Xinting Huang}\tencent,
    \textbf{Tao Fang}\macau,
    \textbf{Wei Bi}\tencent
    \\
    \suda Soochow University \hspace{1cm}
    \tencent Tencent AI Lab  \hspace{1cm}
    \macau University of Macau \\
    \texttt{yzhang21@stu.suda.edu.cn}, \texttt{leyangcui@tencent.com}
}

\begin{document}
\begin{CJK}{UTF8}{gkai}

\maketitle
\renewcommand{\thefootnote}{\fnsymbol{footnote}}
\footnotetext[1]{Work was done during the internship at Tencent AI lab.}
\begin{abstract}
Proprietary Large Language Models (LLMs), such as ChatGPT, have garnered significant attention due to their exceptional capabilities in handling a diverse range of tasks.
Recent studies demonstrate that open-sourced smaller foundational models, such as 7B-size LLaMA, can also display remarkable proficiency in tackling diverse tasks when fine-tuned using instruction-driven data.
In this work, we investigate a practical problem setting where the primary focus is on one or a few particular tasks rather than general-purpose instruction following, and explore whether LLMs can be beneficial and further improved for such targeted scenarios.
We choose the writing-assistant scenario as the testbed, which includes seven writing tasks.
We collect training data for these tasks, reframe them in an instruction-following format, and subsequently refine the LLM, specifically LLaMA, via instruction tuning.
Experimental results show that fine-tuning LLaMA on writing instruction data significantly improves its ability on writing tasks.
We also conduct more experiments and analyses to offer insights for future work on effectively fine-tuning LLaMA for specific scenarios.
Finally, we initiate a discussion regarding the necessity of employing LLMs for only one targeted tasks, taking into account the efforts required for tuning and the resources consumed during deployment.

\end{abstract}
\section{Introduction}
Proprietary Large Language Models (LLMs), such as ChatGPT and GPT4 \cite{openai2023gpt4}, 
have shown exceptional capabilities to follow general human instructions and solve diverse tasks, including but not limited to question answering~\cite{qin2023chatgpt}, machine translation~\cite{jiao2023chatgpt}, information extraction~\cite{wei2023zero}, and grammar correction~\cite{fang2023chatgpt}. Recent studies demonstrate that smaller foundational models, such as LLaMA~\cite{touvron2023llama}, can also display remarkable proficiency in tackling diverse tasks when fine-tuned using instruction-driven data, as exemplified by Alpaca~\cite{alpaca} and Vicuna~\cite{vicuna2023}.  All of these LLMs' accomplishments have represented a significant stride toward the goal of general artificial intelligence.

Due to the great potential of LLMs, people have been considering if LLMs can empower thousands of industries.
 In many practical scenarios, the primary focus of deploying an LLM is on some vertical applications, i.e., one or a few particular tasks. 
For example, Med-PaLM 2 harnesses the power of Google’s LLMs on the medical domain to more accurately and safely answer medical questions~\cite{singhal2023towards}.
In order to facilitate a wide range of NLP tasks in the financial industry, Bloomberg has released its LLM, which has been particularly trained on a vast amount of financial data~\cite{wu2023bloomberggpt}.
Despite the surprising performance of current LLMs for vertical applications, LLMs still often make fatal errors, such as hallucination, or using incorrect knowledge for reasoning in specific domains, making them even inferior to some task-specific supervised-trained small models~\cite{pan2023preliminary,liu2023comprehensive,yuan2023zero}.

Therefore, in this work, we explore whether LLMs can be efficiently improved for certain targeted scenarios.
We choose the writing-assistance scenario as our targeted vertical application. Many existing writing assistants~\cite{fang2023chatgpt,loem2023exploring} pack LLMs as the major backbone models, which can help users improve and refine their texts \cite{shi2022effidit}. 
We carefully select seven representative writing-related tasks,
and collect about 60k training data for these tasks.
We reformulate our data into the instruction-following format, and combine them with some existing instruction-following data for general purposes, such as the data provided by the \textit{Stanford Alpaca} project~\cite{alpaca}, for finetuning LLMs.

After evaluating the performance of fine-tuned LLMs under different settings in the writing assistant scenario, we have several major findings.
First, fine-tuning an LLM to the writing assistant scenario with a few writing instruction data can lead to significant improvement. Especially, smaller fine-tuned LLMs (<10B) have the potential to outperform larger ones (>100B) without finetuning in the targeted vertical domain.
Second, we identify an effective strategy for preserving the general capabilities of finetuned LLMs beyond the targeted tasks, which involves incorporating a mixture of general instruction-following data during the finetuning process. 
We also investigate more factors that may affect the final performance, such as model size, full versus parameter efficient training, etc.
We hope our empirical study can serve as a practical demonstration of the development and deployment of application-specific LLMs. 

We further explore a more focused question: is it necessary to deploy an LLM for the sole purpose of facilitating a single targeted task?
We examine the grammatical error correction (GEC) task as a case study, and compare finetuned LLMs with conventional competitive approaches, considering the performance, training and deployment costs, etc. We observe that fine-tuned LLMs struggle to be comparable to task-specific, lightweight models in GEC, even with considerable additional effort. This may be attributed to some intrinsic limitations of LLMs, such as a higher propensity for generating hallucinations \cite{bang2023multitask, zhang2023siren}. Consequently, we ought to exercise caution and carefully consider the expense before applying LLMs to a specific task.

\section{Related Work}
\paragraph{LLMs for General NLP.} 
The task of language models is to make next token predictions, which may be very simple for modeling and fitting at first glance. However, this is exactly the task that our human beings routinely complete in order to produce sentences that express our thinking and reasoning.
Therefore, a language model, if intelligent enough, should imitate human beings by demonstrating its capacity to deliver its feedback for whatever context/task it receives.
Recent research has shown that scaling language models to larger sizes can lead to significant improvements in performance across a wide spectrum of downstream tasks~\cite{kaplan2020scaling}.
One of the most successful LLMs to date could be OpenAI's GPT series~\cite{brown2020language}, particularly the recently introduced ChatGPT \cite{chatgpt} and GPT4 \cite{openai2023gpt4}.
These LLMs have demonstrated remarkable proficiency in following and executing various user instructions, resulting in widespread adoption by users who regularly interact with these models and even employ them for professional use.

Recent work also revealed that open-sourced foundation LLMs (e.g., LLaMA \cite{touvron2023llama}, BLOOM \cite{scao2022bloom}) fine-tuned on instruction-following demonstrations generated/distilled from powerful proprietary models such as ChatGPT, can show behaviors similar to ChatGPT/GPT4~\cite{peng2023instruction}. Such datasets include Self-Instruct \cite{wang2022self}, Unnatural Instructions \cite{honovich2022unnatural}, Alpaca \cite{alpaca}, GPT4-Alpaca \cite{peng2023instruction}.

\paragraph{LLMs for vertical applications.} %
The immense potential of LLMs has led researchers to explore their applicability across a multitude of industries. 
Numerous attempts have been made to harness LLMs in various vertical domains, such as healthcare \cite{singhal2023towards} and finance \cite{wu2023bloomberggpt}.
Instruction Tuning is a straightforward way to adapt LLMs for a vertical domain, which only needs to prepare the domain-specific datasets presented in natural language descriptions.
For instance, \citet{chung2022scaling} train Flan-PaLM 540B, which is instruction-tuned on 1.8K tasks, and find it outperforms PaLM 540B by a large margin on unseen downstream benchmarks.
Their study also revealed that the diversity of training instructions contributed to improved performance on hold-out tasks. In this work, we demonstrate the importance of both generic and scenario-specific instruction tuning to harness LLMs' capabilities within a constrained scenario.
There are also existing works that perform multi-task instruction tuning of LLMs for specific scenarios, such as machine translation \cite{jiao2023parrot}, information extraction \cite{wang2023instructuie}, and medical QA \cite{wang2023huatuo}. Our work instead focuses on the writing-assistant scenario and dives deeper, studies the best practice of specifying LLMs and discusses the expense.

\section{Specifying LLMs to a Scenario}
\begin{table*}[t]
\centering
\scalebox{0.83}{
\begin{tabular}{lll}
\toprule
\textbf{Task}  & \textbf{Prompt}             & \textbf{Example}                               \\
\midrule
Grammaticality & Fix grammatical errors in the text.      & She went to the \textcolor{red}{\sout{markt}}\textcolor{blue}{market}.                        \\
Fluency        & Make the text more fluent.               & \begin{tabular}[l]{@{}l@{}}\textcolor{red}{\sout{they}}\textcolor{blue}{They} just create \textcolor{blue}{such a good} impression  \textcolor{red}{\sout{such well}}.\end{tabular}          \\
Clarity        & Make the text more concise and readable. & The changes \textcolor{red}{\sout{made}}\textcolor{blue}{improved} the paper \textcolor{red}{\sout{better than before}}. \\
Coherence      & Make the text more cohesive.             & She works hard\textcolor{red}{\sout{. She}}\textcolor{blue}{, therefore she} is successful.             \\
Simplification & Simplify the text.                       & They have poor \textcolor{red}{\sout{visual acuity}}\textcolor{blue}{eyesight}.                  \\
Neutralization & Neutralize the text.                     & Go is \textcolor{blue}{one of} the deepest game in the world.  \\   
Paraphrasing & Paraphrase the text.                     &  \textcolor{red}{\sout{He loves cats the most.}}\textcolor{blue}{Cats are his favorite animals.}  \\   
\bottomrule
\end{tabular}
}
\caption{Illustrations of writing tasks. We use \textcolor{red}{Red} and \textcolor{blue}{Blue} to highlight the deleted and inserted content, respectively.}
\label{tab:ill}
\end{table*}
In this section, we first investigate whether LLMs with fine-tuning are able to achieve superior performance in the writing assistance scenario.
We first carefully select seven writing tasks and ten datasets to create a comprehensive evaluation benchmark.
Then, we collect 60k publicly available training data for six writing tasks and reformulate them in the instruction-following format.
We combine these writing instruction data with about 52k generic instruction data from the \textit{Stanford Alpaca} project \cite{alpaca}. Finally, we fine-tune a cutting-edge open LLM to date, i.e., LLaMA \cite{touvron2023llama}, with them.

\subsection{Benchmark Setting}

Our evaluation benchmark is mainly extended from \textsc{EditEval} \cite{dwivedi2022editeval}\footnote{\url{https://github.com/facebookresearch/EditEval}}, an instruction-based benchmark aiming at measuring models' capability to improve and edit texts.
Specifically, we remove the \textit{Updating} task since this task requires information from external sources, which is not convenient for evaluating. We additionally introduce the \textit{Grammaticality} task.
As shown in Table \ref{tab:ill}, there are seven writing tasks in total, and the details are listed below.

\paragraph{\textbf{Grammaticality}.} This task, often referred to as Grammatical Error Correction (GEC) \cite{bryant2022survey}, aims to rectify to fix the spelling and grammatical errors in given texts. We choose two GEC benchmarks, i.e., CoNLL14 \cite{ng2014conll} and BEA19-dev \cite{bryant2019bea} for our evaluation, which mainly consists of learner essays with expert-annotated corrections.

\paragraph{\textbf{Fluency}.} This task involves correcting errors in a sentence and additionally improving its fluency and naturalness. JFLEG\cite{napoles2017jfleg} is the first dataset that can be used for evaluating fluency. Another dataset, \textsc{IteraTeR} \cite{du2022understanding}, annotates edits from different dimensions, including fluency, clarity, and coherence. We employ the fluency subset of it (ITR-F) here.

\paragraph{\textbf{Clarity}.} The objective of the clarity task is to enhance the conciseness of the text. We use the clarity subset of \textsc{IteraTeR} \cite{du2022understanding} (ITR-L) to evaluate it.

\paragraph{\textbf{Coherence}.} This task focuses on enhancing the cohesiveness of the text. We use the coherence subset of  \textsc{IteraTeR} \cite{du2022understanding} (ITR-O) to evaluate it.

\paragraph{\textbf{Simplification}.} Simplification is the task of making the text simpler without changing its original meaning. The datasets we utilize for simplification are TurkCorpus \cite{xu2016optimizing} and ASSET \cite{alva2020asset}, both have multiple reference simplifications for the original sentence.

\paragraph{\textbf{Neutralization}.} The task of neutralization refers to removing any point of view from texts. To evaluate this, we involve the Wiki Neutrality Corpus (WNC) \cite{pryzant2020automatically}.

\paragraph{\textbf{Paraphrasing}.} Paraphrasing means rewriting a sentence and keeping its meaning unchanged. For paraphrasing, we follow \textsc{EditEval} to use the STS benchmark from the SemEval-2018 shared task \cite{cer2017semeval}.
\begin{table}[!t]
\centering
\scalebox{0.78}{
\begin{tabular}{lllll}
\toprule
 \textbf{Task} & \textbf{Dataset} & \textbf{Abbrev.}  & \textbf{Metric} & \textbf{Size} \\ \bottomrule
 Grammaticality & CoNLL14 & CoNLL & F$_{0.5}$ & 1,312 \\
 Grammaticality & BEA19-dev & BEA & F$_{0.5}$ & 4,384 \\
 Fluency & JFLEG & JFL & GLEU & 747 \\
 Fluency & \textsc{IteraTeR} & ITR-F & SARI & 88 \\
 Clarity & \textsc{IteraTeR} & ITR-L & SARI & 185 \\
 Coherence & \textsc{IteraTeR} & ITR-O & SARI & 35 \\
 Simplification & TurkCorpus & TRK & SARI & 359 \\
 Simplification & ASSET & AST & SARI & 359 \\
 Neutralization & WNC & WNC & SARI & 1,000 \\
 Paraphrasing & STS & STS & SARI & 97 \\
\bottomrule
\end{tabular}
}
\caption{Tasks, datasets, abbreviations used, metrics used, and corresponding test size (number of sentences) in our evaluation benchmark.
} 
\label{tab:bench:data}
\end{table}

\subsection{Evaluation Setting}
The detailed information about our evaluation benchmark is listed in Table \ref{tab:bench:data}.
For each task, we carefully select the corresponding evaluation metrics based on prior work. A brief introduction to these metrics is provided below.

\paragraph{\textbf{Edit-level F$_{0.5}$}.} We first align the source sentence and the hypothesis sentence using the edit-distance algorithm to extract a group of hypothesis edits. Then, we can compare the hypothesis edits with the golden edits to calculate edit-level precision and recall. The F$_{0.5}$ is the harmonic mean of precision and recall, and weight precision twice as recall. We use edit-level F$_{0.5}$ to evaluate the grammaticality task following previous work \cite{dahlmeier2012better, bryant2017automatic}.

\paragraph{\textbf{GLUE}.} This is a variant of BLEU that penalize n-grams changed in the reference but not modified in the hypothesis \cite{napoles2015ground}. This is the official metric of JFLEG, so we choose it to evaluate system performance on JFLEG.

\paragraph{\textbf{SARI}.} For the remaining tasks, we adopt SARI, an n-gram-based metric introduced by \citet{xu2016optimizing} that frequently used for measuring text editing tasks. It computes the arithmetic mean of n-gram F$_1$ for the inserting/deleting/keeping actions, and is proven to correlate closely with human judgments.

\subsection{Instruction Scheme}
Following \citet{alpaca}, we design an instruction scheme to help the model to understand the task.
As shown below, the instruction scheme contains a universal preface, an instruction field to guide task completion, an input field that provides the text to be edited, and a response field that needs LLMs to fill out.

\vspace{-0.1cm}
\begin{table}[!htbp]
\small
    \centering
    \colorbox{gray!8}{
    \begin{tabular}{@{}p{7.3cm}}
    ============ \textsc{Instruction Format} ===========\\\\
    Below is an instruction that describes a task, paired with an input that provides further context. Write a response that appropriately completes the request.\\\\\

    $\#\#\#$Instruction: \\
    \texttt{[Task Prompt]}\\\\
    
    $\#\#\#$Input:  \\
    \texttt{[Input Text]}\\\\
    
    $\#\#\#$Response:  \\
    \texttt{[Output Text]}\\\\
    \end{tabular}}
\end{table}
\vspace{-0.3cm}

The prompts for seven writing tasks are presented in Table \ref{tab:ill}. Currently, we only write a single prompt for each task, which could be potentially problematic. However, in our preliminary experiments, we found that using generic instruction data could enhance the model's ability to understand different semantically similar prompts, which may mitigate this problem.

\begin{table}[!t]
\centering
\scalebox{0.78}{
\begin{tabular}{lll}
\toprule
 \textbf{Task} & \textbf{Dataset} &  \textbf{Size} \\ \bottomrule
 General & Stanford Alpaca &  52,002 \\ \hdashline
 Grammaticality & W\&I+LOCNESS-Train &  10,000 \\ 
 Fluency & \textsc{IteraTeR}-V2-Train &  10,000 \\ 
 Clarity & \textsc{IteraTeR}-V2-Train &  10,000 \\ 
 Coherence & \textsc{IteraTeR}-V2-Train &  10,000 \\ 
 Simplification & TurkCorpus &  2,000 \\ 
 Simplification & Wiki-Large &  8,000 \\ 
  Neutralization & WNC &  10,000 \\ 
\bottomrule
\end{tabular}
}
\caption{The statistics of our training data.
} 
\label{tab:train:data}
\end{table}
\subsection{Training Data}
The overview of the training data involved in our experiments is shown in Table \ref{tab:train:data}.

\paragraph{Generic Instruction Data.} 
This kind of data asks LLMs to solve a wide spectrum of general-purposed tasks, such as writing a poem or listing a travel plan. In practice, we use the \textit{Stanford Alpaca} dataset\footnote{\url{https://github.com/tatsu-lab/stanford_alpaca}}, which is composed of about 52k instruction data generated by querying OpenAI's \texttt{text-davinci-003} API via the self-instruct technique \cite{wang2022self}. Recent studies have demonstrated that fine-tuning LLaMA with such data can effectively enhance its ability to follow human instructions to solve various tasks.

\begin{table*}[t]
\centering
\scalebox{0.7}{
\begin{tabular}{lccccccccccc}
\toprule
                        & \multicolumn{2}{c}{\textbf{Grammaticality}} & \multicolumn{2}{c}{\textbf{Fluency}} & \textbf{Clarity} & \textbf{Cohere.} & \multicolumn{2}{c}{\textbf{Simp.}} & \textbf{Neu.} & \textbf{\begin{tabular}[c]{@{}c@{}}Para. \\ (Hold-out)\end{tabular}} & \textbf{All}        \\ \cmidrule(lr){2-3}\cmidrule(lr){4-5}\cmidrule(lr){6-6}\cmidrule(lr){7-7}\cmidrule(lr){8-9}\cmidrule(lr){10-10}\cmidrule(lr){11-11}\cmidrule(lr){12-12}
\textbf{Model}          & \textbf{CoNLL}         & \textbf{BEA}        & \textbf{JFL}     & \textbf{ITR-F}     & \textbf{ITR-L}   & \textbf{ITR-O}   & \textbf{TRK}     & \textbf{AST}     & \textbf{WNC}  & \textbf{STS}                                                         & \textbf{Avg. Score} \\ \midrule
\textbf{LLaMA-7B}          & 19.21                  & 22.42               & 41.14            & 35.17              & 34.12            & 34.88            & 30.87            & 25.13            & 31.13         & 32.34                                                                & 30.64               \\ \hdashline
\textbf{Alpaca-7B}         & 52.86                  & 36.19               & 59.30            & 48.54              & 35.48            & 37.98            & 40.33            & 43.91            & 34.89         & 41.19                                                                & 43.07               \\
\textbf{\hspace{0.3cm}7B$\rightarrow$13B}         & 53.23                  & 37.52               & 61.05            & 52.01              & 34.70            & 36.17            & 41.15            & 43.02            & 35.71         & 42.99                                                                & 43.76               \\
\textbf{\hspace{0.3cm}GPT3.5$\rightarrow$GPT4 data}        & 50.97                  & 36.31               & 59.13            & 52.35              & 35.10            & 36.31            & 40.22            & 44.84            & 33.24         & 39.64                                                                & 42.81               \\ \hdashline
\textbf{Writing-Alpaca-7B} & 55.88                  & 46.35               & 59.93            & 52.79              & {\bf 39.44}            & 37.12            & 42.67            & 44.57            & 64.43         & {\bf 43.30}                                                                &  48.65               \\
\textbf{\hspace{0.3cm}-- Generic instruction data} & 56.31                  & 46.99               & 60.70            & 56.12              & 38.11            & 36.66            & 40.83            & 42.26            & 63.36         & 27.79                                                                & 46.91               \\
\textbf{\hspace{0.3cm}LoRA$\rightarrow$Full Fine-tuning} & {\bf 56.94}                  & {\bf 47.66}               & {\bf 61.89}           & {\bf 57.35}              & 38.18            & {\bf 43.31}            &     {\bf 42.24}        & {\bf 44.66}            & {\bf 70.95}         & 42.02                                                                & {\bf 50.52}              \\
\bottomrule
\end{tabular}
}
\caption{Main experimental results for seven selected writing tasks. ``Cohere.'' denotes coherence, ``Simp.'' stands for simplification, ``Neu.'' means neutralization, and ``Para.'' denotes paraphrasing. The metric score used for each dataset is illustrated in Table \ref{tab:bench:data}.}
\label{tab:main}
\end{table*}
\paragraph{Writing Instruction Data.} To perform supervised fine-tuning for the writing scenario, we gather 10k training instances for each of the six writing tasks, resulting in a total of 60k training instances. We leave out the paraphrasing task in order to study the zero-shot performance of fine-tuned LLMs when handling general writing tasks.

For the grammaticality task, we randomly select 10k data from the W\&I+LOCNESS dataset~\cite{bryant2019bea}. For fluency, clarity, and coherence, we randomly pick 10k data for each from the fluency, clarity, and coherence subset of the \textsc{IteraTeR}-V2 training set \cite{kim-etal-2022-improving}. 
For simplification, we choose 2k data from the training set of TurkCorpus \cite{xu2016optimizing} and 8k data from the Wiki-Large training set \cite{kauchak2013improving}.
For neutralization, we randomly choose 10k data from the WNC training set \cite{pryzant2020automatically}.
Finally, we re-process all chosen multi-task writing data into the instruction-following format as described in the previous section before training.

\subsection{Implementation Details}
We fine-tune the official version of LLaMA\footnote{\url{https://github.com/facebookresearch/llama}} \cite{touvron2023llama} with the Huggingface Transformers \cite{wolf2020transformers} toolkit.
During training, we optimize LLaMA to output the reference response via cross-entropy loss.
Considering the time and computational resources, we perform parameter-efficient fine-tuning instead of full-model fine-tuning for most of our experiments.
Specifically, we mainly utilize the low-rank adaptation (LoRA) \cite{hu2021lora} technique for computational effectiveness. 
We also compare LoRA with full-model fine-tuning in Section \ref{sec:4.2}.
For the hyper-parameter setting, we basically follow the \textit{Alpaca LoRA} project\footnote{\url{https://github.com/tloen/alpaca-lora}}. All trained models share the same training steps.
All experiments are carried out on 8 Nvidia V100 32GB GPUs.

\subsection{Main Results}
Table \ref{tab:main} shows the main results.
We refer to LLaMA which has been fine-tuned using only generic instruction data as \textbf{Alpaca}, and LLaMA which has been fine-tuned using both generic and writing instruction data as \textbf{Writing-Alpaca}.

\begin{table*}[t]
\centering
\scalebox{0.8}{
\begin{tabular}{lcccccccccc}
\toprule
                        & \multicolumn{2}{c}{\textbf{Grammaticality}} & \multicolumn{2}{c}{\textbf{Fluency}} & \textbf{Clarity} & \textbf{Cohere.} & \multicolumn{2}{c}{\textbf{Simp.}} & \textbf{Neu.} & \textbf{Para.}       \\ \cmidrule(lr){2-3}\cmidrule(lr){4-5}\cmidrule(lr){6-6}\cmidrule(lr){7-7}\cmidrule(lr){8-9}\cmidrule(lr){10-10}\cmidrule(lr){11-11}
\textbf{Model}          & \textbf{CoNLL}         & \textbf{BEA}        & \textbf{JFL}     & \textbf{ITR-F}     & \textbf{ITR-L}   & \textbf{ITR-O}   & \textbf{TRK}     & \textbf{AST}     & \textbf{WNC}  & \textbf{STS}                                                          \\ \midrule
\textbf{OPT-175B$^*$} & --                  & --               & 47.5            & 34.7              & 31.5            & 27.6            & 32.6            & 31.8            & 31.2         & 29.1 \\
\textbf{GPT3-175B$^*$} & --                  & --               & 51.8            & 32.1              & 33.5            & 26.9            & 33.0            & 30.5            & 31.7         & 27.2 \\
\textbf{InstructGPT-175B$^*$} & 51.5        &     40.2            & 59.3           & 48.8              & 35.1            & 35.9            & 38.8            & 38.0            & 35.4         & 42.5  \\
\textbf{ChatGPT} & 53.3                  & 42.7               &   \textbf{62.4}         &      50.9         &     31.5     &      31.0    &      39.9     &   \textbf{47.0}          &      36.3     &  40.9  \\
\textbf{\textit{Writing-Alpaca-7B (ours)}} & \textbf{55.9}                 & \textbf{46.4}               & 59.9            & \textbf{52.8}              & \textbf{39.4}           & \textbf{37.1}           & \textbf{42.7}            & 44.6            & \textbf{64.4}         & \textbf{43.3}    \\ 
\midrule
\textbf{\textit{Prev. SOTA}} & \textit{68.9}                 & \textit{59.6}               & \textit{64.7}           & \textit{52.1}              & \textit{46.2}           & \textit{38.3}           & \textit{38.8}            & \textit{38.0}            & \textit{54.5}         & \textit{42.5}    \\
\bottomrule
\end{tabular}
}
\caption{Comparisons with other instruction-based models. The results for models marked with * are taken from \citet{dwivedi2022editeval}, which may not be directly comparable with our results. The previous SOTA from left to right is achieved by \citet{rothe2021recipe} (CoNLL), \citet{zhang2022syngec} (BEA), \citet{stahlberg2021synthetic} (JFL), \citet{du2022understanding} (ITR-L, ITR-O), \citet{dwivedi2022editeval} (ITR-F, TRK, AST, WNC, STS).}
\label{tab:comp}
\end{table*}

\paragraph{Instruction tuning improves LLaMA's performance on all writing tasks significantly.} Comparing the performance of LLaMA-7B and Alpaca-7B, we can see that fine-tuning foundation language models on a few instruction-following data, even machine-generated, can greatly improve its downstream performance. The average score over seven writing tasks improved from 30.64 to 43.07, which can be attributed to the style or format learned from instruction tuning for interacting with humans.

\paragraph{Scenario-specific instruction tuning can lead to further improvements, but not on all tasks.}
After adding the writing-related instruction data, the overall performance further improves from 43.07 to 48.65 (Alpaca-7B vs. Writing-Alpaca-7B), demonstrating the effectiveness of scenario-specific instruction tuning. The largest gain appears in the neutralization task, where using writing instruction data leads to about 30 points of improvement. 

However, we observe that in certain datasets, the improvement is marginal or even non-existent. For instance, in the JFLEG dataset, Writing-Alpaca surpasses Alpaca by a mere 0.6. 
One possible explanation could be that Alpaca is proficient in producing highly fluent results, thanks to the powerful language modelling capability acquired through unsupervised pre-training on large-scale raw texts.

\paragraph{Adapting LLaMA to writing tasks makes it outperform larger off-the-shelf LLMs while still lag behind some task-specific SOTA.}
As shown in Table \ref{tab:comp}, we compare Writing-Alpaca-7B with other much larger LLMs on our benchmark, such as OPT-175B \cite{zhang2022opt}, GPT3-175B \cite{brown2020language}, InstructGPT-175B (\texttt{text-davinci-001}) \cite{ouyang2022training}, and ChatGPT\footnote{\url{https://chat.openai.com}}. Although our Writing-Alpaca-7B is much smaller, we observe it still outperforms all its counterparts on most writing tasks.
This phenomenon may hint that: adapting a small-size open-sourced LLM to a specific scenario may make it a scenario expert and surpass those much larger off-the-shelf LLMs.
This can be affordable for most companies and individual developers who only need to build targeted applications in practice.

Nonetheless, we observe that Writing-Alpaca-7B sometimes underperforms when compared to previous dataset-specific SOTA. These systems often employ large-scale supervised training data or incorporate modules specially designed for tasks. In Section \ref{sec:task}, we initiate a discussion on the potential application of similar strategies to further enhance the performance of LLMs in a given task.

\subsection{Further Analysis}
\label{sec:4.2}
To gain more insights into specifying LLMs for a targeted scenario, we perform further analysis, as listed below.

\paragraph{Larger LLaMA generally performs better on writing tasks.}
We first try to explore whether larger LLaMA could achieve better performance on our writing benchmark. To this end, we evaluate the performance of Alpaca-13B trained under the same setting as the 7B one (\textbf{7B$\rightarrow$13B} in Table \ref{tab:main}). After increasing the model capacity, Alpaca-13B performs slightly better than its 7B variant (avg. score 43.07$\rightarrow$43.76), showing that larger LLMs can indeed result in improvements.

\paragraph{Fine-tuning LLaMA with better generic instruction data does not necessarily lead to improvements in the writing scenario.}
Recently, \citet{peng2023instruction} attempted to generate 52k generic instruction data using GPT4 \cite{openai2023gpt4}, and claimed that fine-tuning LLaMA with their data leads to much better performance than previous data. Here, we investigate whether better generic instruction data improve performance in our scenario.
As shown in Table \ref{tab:main} (\textbf{GPT3.5$\rightarrow$GPT4 data}), we can see the answer is no (avg. score 43.07 $\rightarrow$ 42.81).
We think that better generic data could further enhance LLaMA's general ability to solve various open-ended tasks and interact with users, but it may not improve LLaMA's specific ability in a constrained scenario.
One possible explanation could be that generic instruction data only help LLMs align with human preference, and expose the knowledge that was already acquired from pre-training, whereas it does not teach LLMs any new scenario-specific knowledge \cite{zhou2023lima}.

\begin{table}[!t]
\centering
\scalebox{0.7}{
\begin{tabular}{ll}
\toprule
 \rowcolor[gray]{.9}\textbf{User Input} & \begin{tabular}[l]{@{}l@{}} \textit{Paraphrase the text:} A man sitting \\ on a couch and watching television. \end{tabular} \\\hline
 \textbf{\begin{tabular}[l]{@{}l@{}}System Response \\ \hspace{0.3cm} w/o. generic instruction\end{tabular}} &  \begin{tabular}[l]{@{}l@{}} A man is sitting on a couch and \\ watching television. \end{tabular} \\ \hline
 \textbf{\begin{tabular}[l]{@{}l@{}}System Response \\ \hspace{0.3cm} w/. generic instruction\end{tabular}} & \begin{tabular}[l]{@{}l@{}} Someone is relaxing on a couch \\ in front of the television.\end{tabular} \\
 \rowcolor[gray]{.9}\textbf{User Input} & Tell me about panda. \\\hline
 \textbf{\begin{tabular}[l]{@{}l@{}}System Response \\ \hspace{0.3cm} w/o. generic instruction\end{tabular}} & Tell me about \textcolor{red}{pandas}. \\ \hline
 \textbf{\begin{tabular}[l]{@{}l@{}}System Response \\ \hspace{0.3cm} w/. generic instruction\end{tabular}} & \begin{tabular}[l]{@{}l@{}}Pandas are large mammals native \\  to China and Southeast Asia. \\ They are known for their distinct \\  black-and-white coloring and \\  their diet of bamboo. They are \\  endangered due to habitat loss \\ and poaching.\end{tabular} \\ 
\bottomrule
\end{tabular}
}
\caption{A case study of how generic instruction tuning keeps the generic ability, e.g., paraphrase, chat.
} 
\label{tab:case}
\end{table}

\begin{table*}[!t]
\centering
\scalebox{0.84}{
\begin{tabular}{lccccccccc}
\toprule
\multirow{2}{*}{\textbf{Model}} & \multirow{2}{*}{\textbf{\#Data}}  & \multicolumn{3}{c}{\textbf{CoNLL}} & \multicolumn{3}{c}{\textbf{BEA}} & \multicolumn{1}{c}{\textbf{Training Speed}} & \multicolumn{1}{c}{\textbf{Inference Speed}}
\\ \cmidrule(lr){3-5}\cmidrule(lr){6-8}\cmidrule(lr){9-9}\cmidrule(lr){10-10}
                         & & \textbf{P}      & \textbf{R}     & \textbf{F$_{0.5}$}     & \textbf{P}     & \textbf{R}     & \textbf{F$_{0.5}$} & \textbf{GPU hours} &  \textbf{Instances / second} \\ \midrule
 \textbf{RoBERTa-Large} & 2.4M & 75.6 & 44.5 & 66.3	& 66.0 & 33.8 & 55.5 & 21 & 274.7 \\
 \textbf{T5-Large} &  2.4M  & 72.2 & 51.4 & 66.8 & 60.5 & 43.1 & 56.0 & 75 & 25.3\\ \hdashline
 \textbf{Writing-Alpaca-7B} & 11.2k  & 68.0 & 32.7 & 55.9 & 53.5 & 30.3 & 46.4 & 41 & 2.1\\
 \textbf{LLaMA-7B-GEC} &  2.4M & 70.3 & 50.7 & 65.2 & 58.5 & 43.1 & 54.6 & 191  & 2.1\\
 \textbf{LLaMA-13B-GEC} & 2.4M  & 72.7 & 51.1 & 67.0 & 60.9 & 42.6 & 56.1 & 397 & 0.7 \\
\bottomrule
\end{tabular}
}

\caption{Performance and resource cost comparison of different models trained for the grammaticality task.
} 
\label{tab:clang8}
\end{table*}

\paragraph{Generic instruction data is important to maintain the generalization ability of LLaMA.}
To highlight the importance of generic instruction data, we further conduct an experiment that only uses writing instruction data to fine-tune LLaMA (\textbf{-- Generic instruction data} in Table \ref{tab:main}).
The performance drops from 48.65 to 46.91 after excluding the generic instruction data. Specifically, the model degenerates heavily in the hold-out paraphrasing task that lacks supervised data (43.30$\rightarrow$27.79). This suggests that generic instruction data plays a key role in activating LLaMA's generalization ability to handle general writing tasks.
As observed from the first case in Table \ref{tab:case}, when generic instruction tuning is not employed, the model tends to fall short in paraphrasing the text, focusing solely on GEC. However, this issue was addressed with the addition of generic instruction data.

Another phenomenon is that without generic instruction data, the model seems only to edit all texts entered by users and fails to understand their true intentions. Another case is provided in Table \ref{tab:case}. Even though our objective is to specialize LLMs towards a specific scenario, we still prefer to preserve their generic capabilities. Consequently, the generic instruction data is indispensable.

\paragraph{Full-model fine-tuning leads to better performance than LoRA on our benchmark.}
We perform parameter-efficient training using LoRA \cite{hu2021lora} for most of our experiments due to the constraint of time and
computational resources. To figure out the effects of only tuning a small part of the parameters, we compare LoRA with full-model fine-tuning in Table \ref{tab:main} (\textbf{LoRA$\rightarrow$Full Fine-tuning}). We keep the main hyperparameters of these two training methods consistent to make a fair comparison. We can observe that fine-tuning the full parameters of LLaMA leads to much better performance than using LoRA in our writing-related tasks (avg. score 48.65 $\rightarrow$ 50.52). However, it is worth noting that fine-tuning the full model typically requires about five times more training hours.

\section{Specifying LLMs to a Task}
\label{sec:task}
Table~\ref{tab:comp} demonstrates that Writing-Alpaca outperforms larger off-the-shelf LLM counterparts in the writing assistant scenario. However, when compared to previous SOTA systems specifically designed for individual tasks, Writing-Alpaca exhibits mixed results. This observation prompts an intriguing research question: can LLMs surpass conventional SOTA on specific tasks if we dedicate adequate effort? To address this question, we select the grammaticality task, where a noticeable performance gap exists between Writing-Alpaca and the previous SOTA, as the focus of our experiments.

\subsection{Main Results}

We select two conventional models that achieved SOTA performance in the grammaticality task for comparison: 1) RoBERTa-Large \cite{liu2019RoBERTa} based on the GECToR framework \cite{omelianchuk2020gector} which comprises 354M parameters, and 2) T5-Large \cite{raffel2020exploring}, a pre-trained seq2seq model containing 770M parameters. 
Both models are fine-tuned on the CLang8 GEC training data \cite{rothe2021recipe}, which consists of approximately 2.4M high-quality sentence pairs.

We first aim to elicit LLaMA's ability by fine-tuning it with large-scale, high-quality training data for a particular task. Consequently, we conduct an experiment in which LLaMA-7B is finetuned on the CLang8 data (referred to as \textbf{LLaMA-7B-GEC}), following the practice of previous SOTA methods in this field. Table \ref{tab:clang8} reveals that, when fine-tuned with substantial in-domain training data, LLaMA-7B-GEC significantly outperforms Writing-Alpaca on both CoNLL and BEA benchmarks. 
This finding indicates that increasing the volume of task-specific training data is also advantageous for LLMs, as it is for conventional smaller models.

We subsequently explore whether increasing model capacity (scaling LLaMA from 7B to 13B, referred to as \textbf{LLaMA-13B-GEC}) could enhance the performance of LLaMA in the grammaticality task. 
As observed, LLaMA-13B-GEC exhibits a slight improvement in performance compared to LLaMA-7B-GEC. 
In contrast, as reported in \citet{rothe2021recipe}, T5-large (770M) shows near 3 F$_{0.5}$ increase compared to T5-base (220M) on CoNLL. 
We speculate that this may be because the 7B-size LLaMA has already learned most of the knowledge required for the grammaticality task.

\pgfdeclarepatternformonly{dense crosshatch dots}{\pgfqpoint{-1pt}{-1pt}}{\pgfqpoint{5pt}{5pt}}{\pgfqpoint{1.5pt}{1.5pt}}%
{
    \pgfpathcircle{\pgfqpoint{0pt}{0pt}}{.5pt}
    \pgfpathcircle{\pgfqpoint{3pt}{3pt}}{.5pt}
    \pgfusepath{fill}
}

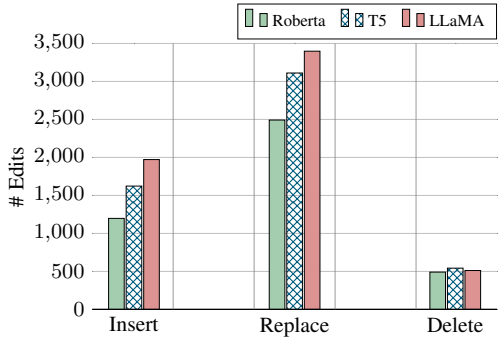
\begin{figure}[tb]
  \small
  \centering
  \scalebox{0.9}{
    \begin{tikzpicture}[
      trim axis left,
      trim axis right,
      my grid/.style={thin, gray, opacity=0.8},
    ]
    \pgfplotstableread{wer.dat}\loadedtable
    \begin{axis}[
        footnotesize,
        ymajorgrids,
        yminorgrids,
        tick align=inside,
        axis line style={opacity=0},
        tickwidth=0pt,
        width=7.5cm, height=5.5cm,
        enlarge x limits=0.13,
        ybar=2*\pgflinewidth,
        bar width=6.5pt,
        legend style={at={(1, 1.15)},
            anchor=north east, legend columns=-1,
            /tikz/every even column/.append style={column sep=0.15cm}},
        symbolic x coords={
            Insert,
            Replace,
            Delete
          }, %
        xtick=data,
        x tick label style={font=\small, align=center}, %
        xtick distance=1,
        scaled ticks=false,
        ymin=0, ymax=3500,
        ytick={0, 500,1000,1500,2000,2500,3000,3500},
        yticklabel={\pgfmathparse{\tick}\pgfmathprintnumber[fixed]{\pgfmathresult}}, %
        ylabel={\small \# Edits},
        ylabel style={at={(0.02,0.5)}}
      ]
      \addplot[draw=black, fill=forestgreen!50] table[x=category,y=Roberta] {\loadedtable};
      \addlegendentry{\scriptsize Roberta}
      \addplot[draw=black, pattern=crosshatch, pattern color=midnightblue] table[x=category,y=T5] {\loadedtable};
      \addlegendentry{\scriptsize T5}
      \addplot[draw=black, fill=brickred!50] table[x=category,y=LLaMA] {\loadedtable};
      \addlegendentry{\scriptsize LLaMA}
      \draw [semithick, black] (axis description cs:0,0) -- (axis description cs:1,0); %
      \draw [my grid] (axis description cs:0.2,0) -- (axis description cs:0.2,1);
      \draw [my grid] (axis description cs:0.4,0) -- (axis description cs:0.4,1);
      \draw [my grid] (axis description cs:0.6,0) -- (axis description cs:0.6,1);
      \draw [my grid] (axis description cs:0.8,0) -- (axis description cs:0.8,1);
    \end{axis}
  \end{tikzpicture}
  }

  \caption{
    The number of correction edits predicted by each model, which is decomposed by the edit action.
  }
  \label{fig:wer}
\end{figure}

\subsection{Controversy of task-specific LLMs}

\paragraph{Making LLaMA comparable to advanced small models in the grammaticality task requires expensive expenses.}
Upon simultaneously increasing the amount of task-specific training data and the model size, LLaMA-13B-GEC ultimately achieves performance comparable to that of previous smaller, task-specific models in the grammaticality task. However, we argue the associated additional costs should not be overlooked.

First, we consider the increased training costs. We measure the GPU hours consumed during training for each model, as detailed in Table~\ref{tab:clang8}. All models are trained using the Huggingface Transformers toolkit and 8 Nvidia V100 32GB GPUs. LLaMA-13B-GEC requires approximately 397 GPU hours for training, which is considerably more time-consuming than RoBERTa-Large (18.9$\times$) and T5-Large (5.3$\times$). It is worth noting that we have already employed the LoRA technique when training LLaMA-13B-GEC to reduce costs. Utilizing full fine-tuning would take about five times longer.

Second, since GEC models are frequently deployed in online services, inference speed is also a critical metric in the grammaticality task. Although LLaMA-13B-GEC achieves similar performance to T5-Large and RoBERTa-Large, its inference speed is much slower. With the same beam size of 16, LLaMA-13B-GEC only predicts 0.7 samples per second, while RoBERTa-Large and T5-Large can predict 274.7 and 25.3 samples, respectively.

\paragraph{Hallucination may be one major issue that prevents LLaMA from surpassing small models in the grammaticality task.}
Although LLMs encode abundant knowledge via extensive pre-training, there exists a ``memory distortion'' problem when they mobilize such knowledge, and hence hallucination occurs \cite{peng2023check}. Hallucination refers to the generation of nonfactual, untruthful information. LLMs are demonstrated to display more serious hallucinations compared with smaller LMs \cite{bang2023multitask, zhang2023siren}.
After carefully examining the results of different systems, we conclude that one main stumbling block that prevents LLMs from significantly surpassing small models in the grammaticality task could be hallucinations.
First, we analyze the action type of correction edits from each model in Figure~\ref{fig:wer}.
Compared to smaller models, LLaMA generates more insertion and replacement edits. 
These edits are more prone to introduce factually incorrect or irrelevant information.
Second, we count the percentage of corrections involving entities using spaCy. 9.7\% of corrections produced by LLaMA involve entities, while the figures for RoBERTa and T5 are only 2.5\% and 3.3\%, respectively.
Table~\ref{tab:case} provides an example case where LLaMA mistakenly alters a person's name (Ashby$\Rightarrow$Kate Ashby), while T5 performs well.

\paragraph{Summary.} Overall, before deploying LLMs for a specific task, we recommend users meticulously assess whether the additional overhead introduced is acceptable. Meanwhile, LLMs may not always be a panacea for solving specific tasks. We should carefully consider the characteristics of downstream tasks and the potential limitations of LLMs.

\begin{table}[!t]
\centering
\scalebox{0.68}{
\begin{tabular}{ll}
\toprule
 \textbf{Input} & Dear Mrs Ashby, My name is Andrea Cocci. \\ \hline
 \textbf{T5 Output} & Dear Mrs Ashby, \textcolor{seagreen}{my} name is Andrea Cocci. \\
 \textbf{LLaMA Output} & Dear Mrs \textcolor{red}{Kate} Ashby, \textcolor{seagreen}{my} name is Andrea Cocci. \\
\bottomrule
\end{tabular}
}
\caption{A case study of the hallucination problem of LLaMA in GEC. We use \textcolor{seagreen}{Green} and \textcolor{red}{Red}  to highlight the correct and erroneous modifications, respectively.
} 
\label{tab:case}
\end{table}
\section{Conclusion}
In this study, we investigated the customization of LLMs for vertical applications. We conducted experiments using LLaMA and primarily focused on the writing-assistant scenario, which encompasses seven writing-related tasks. Experimental results demonstrate that scenario-specific instruction tuning enables LLMs to significantly enhance their performance in a targeted scenario, surpassing larger general-purpose LLM counterparts. 
Furthermore, we delved into the necessity and expense associated with employing LLMs for only one targeted task, taking GEC as a case study.
\section*{Limitations}
In this study, we focus on one type of LLM, specifically LLaMA. Further experiments involving other open-source LLMs, such as BLOOM~\cite{scao2022bloom} and Falcon~\cite{penedo2023refinedweb}, are necessary to substantiate the generalizability of our findings. Additionally, our primary focus is on a single scenario and task, namely, writing assistance and GEC. 
We plan to expand our research to encompass a broader range of scenarios and tasks, thereby providing more comprehensive insights.
\section*{Ethics Statement}
All experimental data utilized in this study are publicly available, and our use of them aligns with their intended purposes. We download and use LLaMA in our research after obtaining permission.

\bibliography{anthology,custom}
\bibliographystyle{acl_natbib}

\end{CJK}
\end{document}